%% file: root.tex
\documentclass[letterpaper, 10 pt, conference]{ieeeconf} 

\IEEEoverridecommandlockouts                        

\usepackage{graphics}
\usepackage{amsmath} 
\usepackage{amssymb}
\usepackage{cite}
\usepackage{tikz}
\usepackage{amsmath}
\usepackage{tabularx}
\usepackage{array}
\usepackage{booktabs}
\usepackage{bm}
\usepackage{algorithm}
\usepackage{algorithmic}
\usepackage{bbm}
\usepackage{adjustbox}
\usepackage{caption}

\usepackage{tablefootnote}

\usepackage{hyperref}

\definecolor{green}{RGB}{11,155,13}

\DeclareMathOperator*{\argmin}{argmin}

\title{\LARGE \bf
Learning to Model and Plan for Wheeled Mobility on \\Vertically Challenging Terrain
}

\author{Aniket Datar, Chenhui Pan, and Xuesu Xiao
\thanks{All authors are with the Department of Computer Science, George Mason University {\tt\scriptsize \{adatar, cpan7, xiao\}@gmu.edu}}
}

\begin{document}
\maketitle
\thispagestyle{empty}
\pagestyle{empty}

\input{content/abstract.tex}

\input{content/intro}
\input{content/related}
\input{content/approach}
\input{content/implementation}
\input{content/experiment}
\input{content/conclusion}

\bibliographystyle{IEEEtran}
\bibliography{IEEEabrv,references}
\end{document}

%% file: content/abstract.tex
\begin{abstract}
Most autonomous navigation systems assume wheeled robots are rigid bodies and their 2D planar workspaces can be divided into free spaces and obstacles. However, recent wheeled mobility research, showing that wheeled platforms have the potential of moving over vertically challenging terrain (e.g., rocky outcroppings, rugged boulders, and fallen tree trunks), invalidate both assumptions. Navigating off-road vehicle chassis with long suspension travel and low tire pressure in places where the boundary between obstacles and free spaces is blurry requires precise 3D modeling of the interaction between the chassis and the terrain, which is complicated by suspension and tire deformation, varying tire-terrain friction, vehicle weight distribution and momentum, etc. In this paper, we present a learning approach to model wheeled mobility, i.e., in terms of vehicle-terrain forward dynamics, and plan feasible, stable, and efficient motion to drive over vertically challenging terrain without rolling over or getting stuck. We present physical experiments on two wheeled robots and show that planning using our learned model can achieve up to 60\% improvement in navigation success rate and 46\% reduction in unstable chassis roll and pitch angles. 
\end{abstract}

%% file: content/intro.tex
\section{INTRODUCTION}
\label{sec::intro}
Wheeled robots, arguably the most commonly used mobile robot type, have autonomously moved from one point to another in a collision-free and efficient manner in the real world, e.g., transporting materials in factories or warehouses~\cite{kiva},  vacuuming our homes or offices~\cite{irobot}, and delivering food or packages on sidewalks~\cite{scout}. Thanks to their simple motion mechanism, most wheeled robots are treated as rigid bodies moving through planar workspaces. After tessellating their 2D workspaces into obstacles and free spaces, classical planning algorithms plan feasible paths in the free spaces that are free of collisions with the obstacles~\cite{fox1997dynamic, quinlan1993elastic, rosmann2017kinodynamic, xiao2022autonomous, perille2020benchmarking, nair2022dynabarn}. 

\begin{figure}[t]
    \centering
    \vspace{+6pt}
    \includegraphics[width=1\columnwidth]{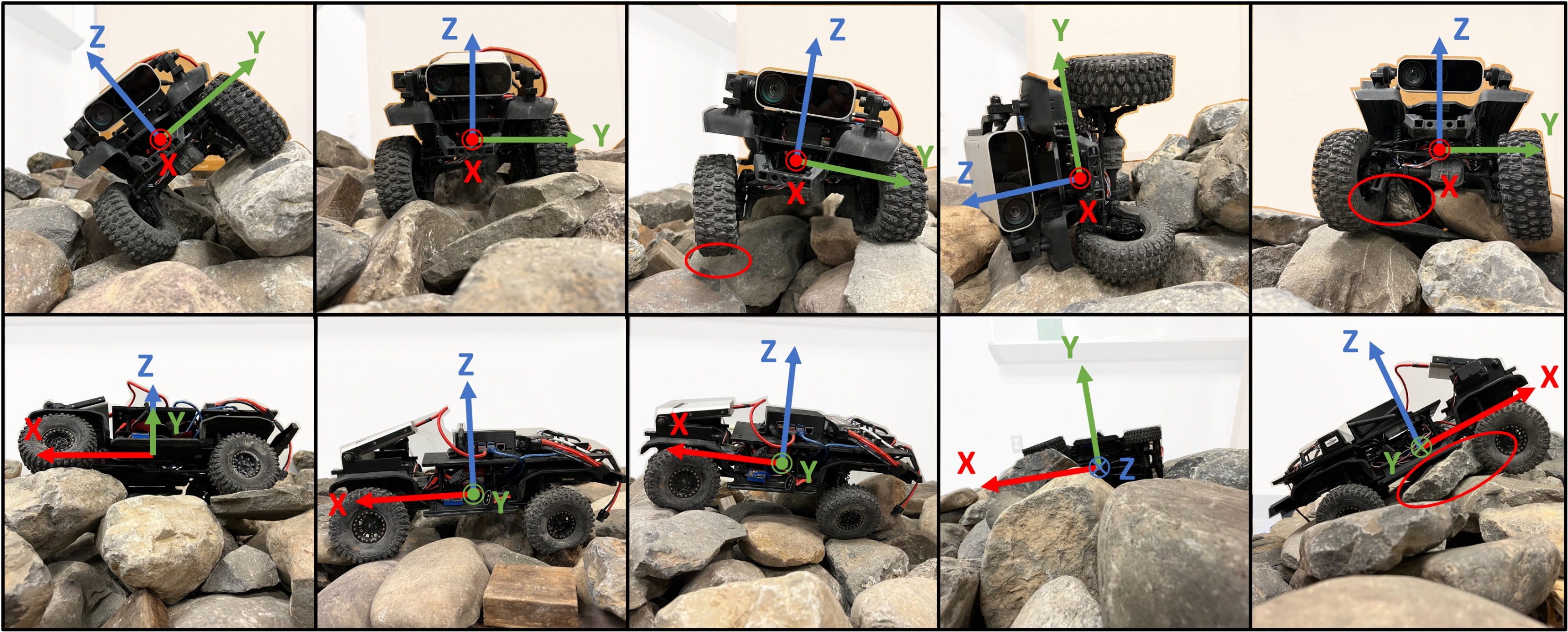}
    \caption{Front and side view (1st and 2nd row) of a wheeled robot navigating vertically challenging terrain: (from left to right) large roll angle, stable chassis, suspended wheel, roll-over, and get-stuck. }
    \label{fig::overview}
\end{figure}

However, recent advances in wheeled mobility have shown that even conventional wheeled robots (i.e., without extensive hardware modification such as active suspensions~\cite{cordes2014active, islam2017novel, jiang2019lateral} or adhesive materials~\cite{liu2018anyclimb}) have previously unrealized potential to move over vertically challenging terrain (e.g., in mountain passes with large boulders or dense forests with fallen trees)~\cite{murphy2014disaster, xiao2018review, mcgarey2016system}, where vehicle motion is no longer constrained to a 2D plane \cite{datar2023toward} (Fig.~\ref{fig::overview}). In those environments, neither assumptions of rigid vehicle chassis and clear delineation between obstacles and free spaces in a simple 2D plane are valid~\cite{xiao2015locomotive, murphy2016two, xiao2017uav, xiao2021autonomous}. Thanks to the long suspension travel and reduced tire pressure, off-road vehicle chassis are able to drive \emph{over} obstacles (rather than to \emph{avoid} them) and experience significant deformation to conform with the irregular terrain underneath the robot, which will be otherwise deemed as non-traversable according to conventional navigation systems. Therefore, autonomously navigating wheeled robots in vertically challenging terrain without rolling over or getting stuck requires a precise understanding of the 3D vehicle-terrain interaction. 

In this paper, we develop a learning approach to model 3D vehicle-terrain interactions and plan vehicle trajectories to drive wheeled robots on vertically challenging terrain. Considering the difficulty in analytically modeling and computing vehicle poses using complex vehicle dynamics~\cite{jazar2008vehicle, yan2019analysis, aly2013vehicle} in real time, we adopt a data-driven approach to model the forward vehicle-terrain dynamics based on terrain elevation maps along potential future trajectories. We develop a Wheeled Mobility on Vertically Challenging Terrain (\textsc{wm-vct}) planner, which uses our learned model's output in a novel cost function in 3D and produces feasible, stable, and efficient motion plans to autonomously navigate wheeled robots on vertically challenging terrain. We present extensive physical experiment results on two wheeled robot platforms and compare our learning approach against four existing baselines and show that our learned model can achieve up to 60\% improvement in navigation success rate and 46\% reduction in unstable chassis roll and pitch angles.

%% file: content/related.tex
\section{RELATED WORK}
\label{sec::related}
We review related work in 2D robot navigation in planar workspaces, facilitating mobility on vertically challenging terrain, and machine learning approaches for mobile robots. 

\subsection{2D Rigid-Body Navigation in Planar Workspaces}
Roboticists have been developing classical ground navigation planners for decades~\cite{fox1997dynamic, quinlan1993elastic, rosmann2017kinodynamic}. Recently, researchers have also investigated high-speed~\cite{atreya2022high, xiao2021learning}, off-road~\cite{manduchi2005obstacle, jackel2006darpa, mousazadeh2013technical}, and social~\cite{mirsky2021conflict, karnan2022socially, chen2017socially, xiao2022learning, francis2023principles, park2023learning, nguyen2023toward} navigation. In the aforementioned research thrusts, most robots are modeled as 2D rigid bodies (e.g., bounding boxes) and most feasible navigation plans are in 2D and collision-free, regardless of the robot type (e.g., wheeled or tracked). 
In order to allow wheeled mobile robots to venture into other difficult-to-reach spaces, recent work has extended wheeled mobility to vertically challenging terrain~\cite{datar2023toward}, considering that vertical protrusions from the ground are not uncommon in real-world unstructured environments~\cite{murphy2014disaster, xiao2018review}. However, both rigid body and planar workspace assumptions are no longer valid in such spaces, requiring new methods to model the interactions between the non-rigid robots and 3D environments. 

\subsection{Robot Mobility on Vertically Challenging Terrain}
Most research aiming at allowing robots to move in vertically challenging environments are from the hardware side. Still treating vehicles as rigid bodies, tracked vehicles are expected to crawl over more rugged terrain than wheeled platforms due to the increased surface contact and therefore propulsion~\cite{vu2008autonomous}, while adhesive materials~\cite{liu2018anyclimb} and tethers~\cite{mcgarey2016system} allow robots to overcome gravity while climbing vertical slopes. Relaxing the assumption of rigid body, vehicles with active suspensions have been developed~\cite{cordes2014active, islam2017novel, jiang2019lateral} to proactively maintain a stable pose of the chassis on vertically challenging terrain. Highly articulated systems, e.g., legged~\cite{Fankhauser2018ProbabilisticTerrainMapping, kumar2021rma}, wheel-legged~\cite{zheng2022mathbf, xu2023whole}, or snake~\cite{wright2007design, xiao2015locomotive} robots,  are another choice to negotiate through such terrain with a stable torso, or virtual chassis for limbless snake robots~\cite{rollinson2011virtual}, by solving many Degrees-of-Freedom (DoFs) of the robot joints. However, despite the versatility to overcome verticality, such specialized hardware are expensive, inefficient (especially on flat terrain), and not as common as conventional wheeled robots.

\subsection{Machine Learning for Robot Mobility}
In addition to the aforementioned classical methods, roboticists have also started utilizing data-driven approaches for robot mobility~\cite{xiao2022motion}. Learning from data, robots no longer need hand-crafted models~\cite{xiao2021learning, sivaprakasam2021improving, karnan2022vi}, cost functions~\cite{vasquez2014inverse, wigness2018robot, xiao2022learning, sikand2022visual}, or planner parameters~\cite{xiao2020appld, wang2021appli, wang2021apple, xu2021applr, xiao2022appl, xu2021machine}, see beyond sensor range  \cite{everett2019planning, meng2023terrainnet}, and acquire navigation behaviors end-to-end from sensor data~\cite{pan2020imitation, faust2018prm, pfeiffer2017perception, xiao2021toward, xiao2021agile, wang2021agile, liu2021lifelong, xu2023benchmarking, karnan2022voila}. Researchers have also used end-to-end Behavior Cloning (BC)~\cite{bojarski2016end} to address wheeled mobility on vertically challenging terrain~\cite{datar2023toward}. Considering the power of machine learning and drawbacks of end-to-end BC (e.g., data-hungry, prone to overfitting, and not generalizable), this work takes a structured learning approach to only model the vehicle-terrain forward dynamics, then constructs a novel cost function in 3D specifically tailored for vertically challenging terrain,  and finally plans trajectories to move robots toward their goal. 

%% file: content/approach.tex
\section{APPROACH}
\label{sec::approach}
The difficulties in navigating a wheeled mobile robot on vertically challenging terrain are two fold: (1) the high variability of vehicle poses due to the irregular terrain underneath the robot may overturn the vehicle (rolling-over, 4th column in Fig.~\ref{fig::overview}); (2) not being able to identify that a certain terrain patch is beyond the robot's mechanical limit and therefore needs to be circumvented may get the robot stuck (immobilization, 5th column in Fig.~\ref{fig::overview}). Therefore, this work takes a structured learning approach to address both challenges by learning a vehicle-terrain forward dynamics model based on the vertically challenging terrain underneath the vehicle, using it to rollout sampled receding-horizon trajectories, and minimizing a cost function to reduce the chance of rolling-over and immobilization and to move the vehicle toward the goal. 

\subsection{Motion Planning Problem Formulation}
\begin{figure*}[ht]
\centering
\includegraphics[width=0.5\textwidth]{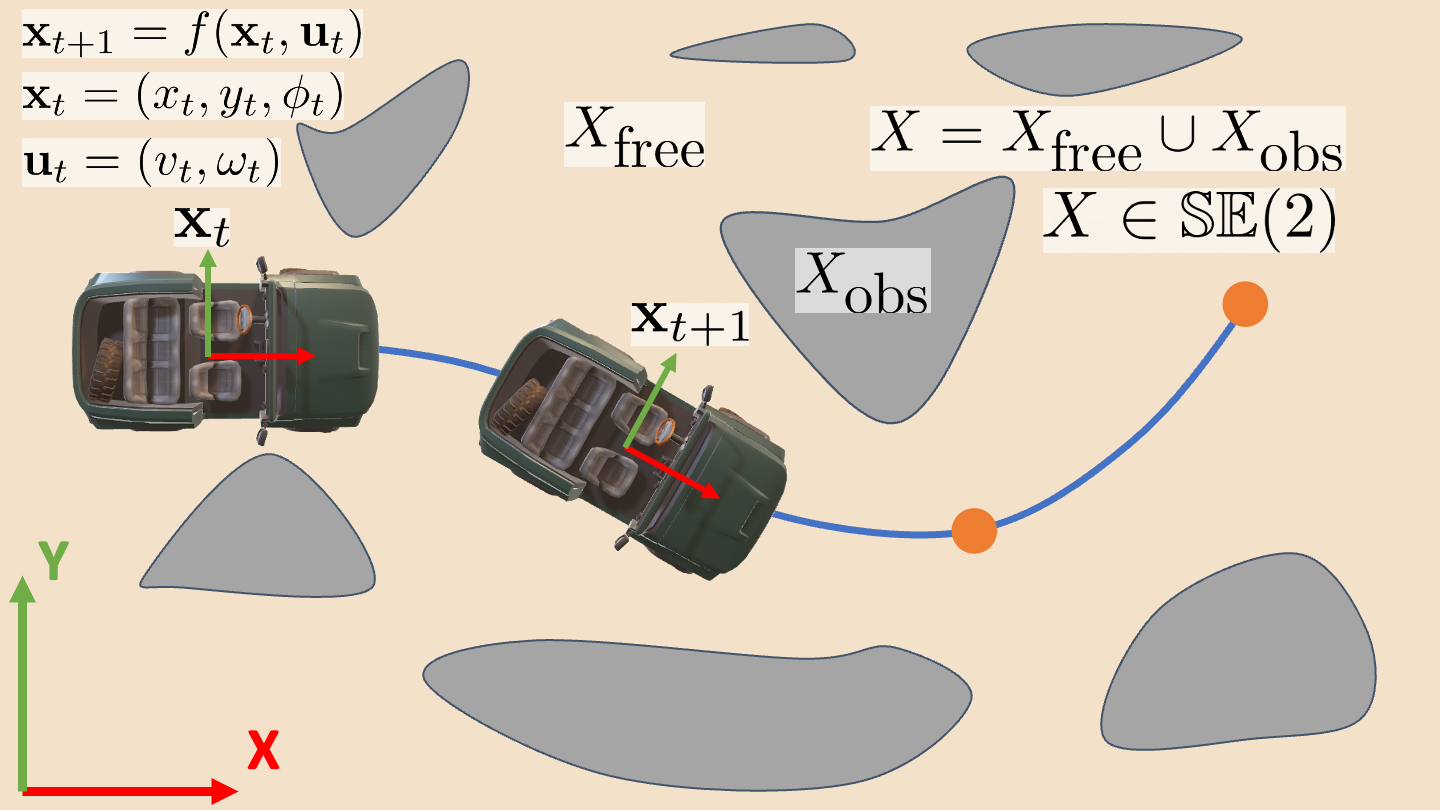}%
\includegraphics[width=0.5\textwidth]{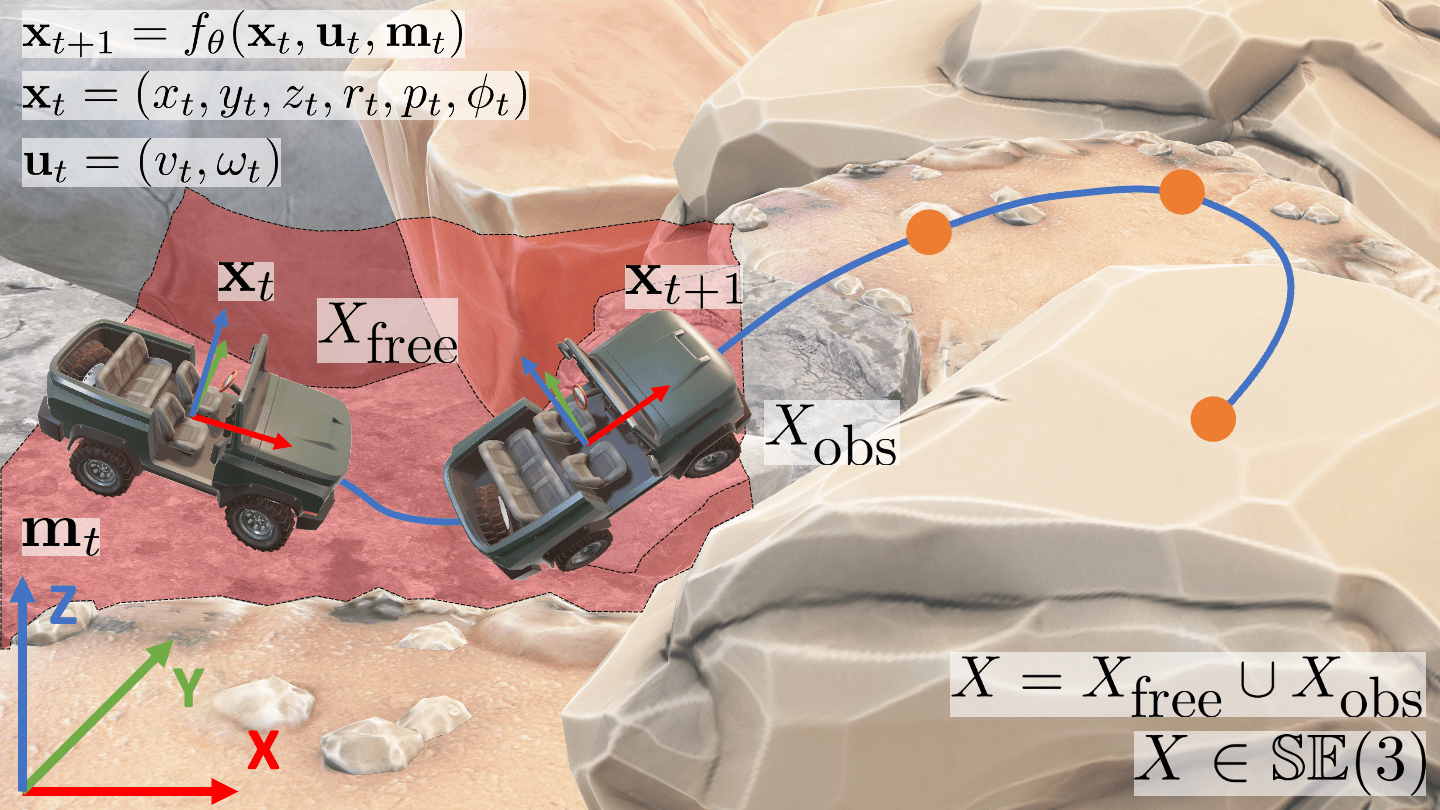}%
\caption{2D navigation in $\mathbb{SE}(2)$ vs. 3D, 6-DoF navigation on vertically challenging terrain in $\mathbb{SE}(3)$.}
\label{fig::comparison}
\end{figure*}

Consider a discrete vehicle dynamics model of the form $\mathbf{x}_{t+1}=f(\mathbf{x}_t, \mathbf{u}_t)$, where $\mathbf{x}_t \in X$ and $\mathbf{u}_t\in U$ denote the state and input space respectively. In the normal case of 2D navigation planning (Fig.~\ref{fig::comparison} left), $X \subset \mathbb{SE}(2)$ and $X=X_\textrm{free} \cup X_\textrm{obs}$, where $X_\textrm{free}$ and $X_\textrm{obs}$ denote free spaces and obstacle regions. $\mathbf{x}_t$ includes the translations along the $\mathbf{x}$ and $\mathbf{y}$ axis ($x$ and $y$) and the rotation along the $\mathbf{z} = \mathbf{x}\times \mathbf{y}$ axis (yaw) of a fixed global coordinate system. For input, $\mathbf{u}_t = (v_t, \omega_t) \in U \subset \mathbb{R}^2$, where $v_t$ and $\omega_t$ are the linear and angular velocity. Finally, let $X_\textrm{goal}\subset X$ denote the goal region. The motion planning problem for the conventional 2D navigation case is to find a control function $u: \{t\}_{t=0}^{T-1} \rightarrow U$ that produces an optimal path $\mathbf{x}_t \in X_\textrm{free}, \forall t\in \{t\}_{t=0}^T$ from an initial state $\mathbf{x}_0=\mathbf{x}_\textrm{init}$ to the goal region $\mathbf{x}_T\in X_\textrm{goal}$ that follows the system dynamics $f(\cdot, \cdot)$ and minimizes a given cost function $c(x)$, which maps from a state trajectory $x: \{t\}_{t=0}^T \rightarrow X$ to a positive real number. In many cases, $c(x)$ is simply the total time step $T$ to reach the goal. Considering the difficulty in finding the absolute minimal-cost state trajectory, many mobile robots use sampling-based motion planners to find near-optimal solutions~\cite{karaman2010incremental, karaman2011anytime}.

Conversely, in our case of wheeled mobility on vertically challenging terrain, vehicle state $X \subset \mathbb{SE}(3)$ (i.e., translations and rotations along the $\mathbf{x}$, $\mathbf{y}$, and $\mathbf{z}$ axis) with the same input $\mathbf{u}_t = (v_t, \omega_t) \in U \subset \mathbb{R}^2$. The system dynamics enforces that $\mathbf{x}_t$ is always ``on top of'' a subset of $X_\textrm{obs}$ (i.e., vertically challenging terrain underneath and supporting the robot) or some boundary of $X$ (i.e., on a flat ground) due to gravity, requiring a 3D, 6-DoF vehicle-terrain dynamics model in $\mathbb{SE}(3)$ (Fig. \ref{fig::comparison} right). 

\subsection{Vehicle-Terrain Dynamics Model Learning}
Compared to the simple 2D vehicle dynamics in $\mathbb{SE}(2)$, our non-rigid vehicle-terrain dynamics on vertically challenging terrain in $\mathbb{SE}(3)$ becomes more difficult to model, considering the complex interaction between the terrain and chassis via the long suspension travel and deflated tire pressure of off-road vehicles to assure adaptivity and traction (Fig.~\ref{fig::overview}). Therefore, this work adopts a data-driven approach to learn the vehicle-terrain dynamics model, which can be used to rollout trajectories for subsequent planning. 

To be specific, $\mathbf{x}_t = (x_t, y_t, z_t, r_t, p_t, \phi_t)$, where the first and last three denote the translational ($x$, $y$, $z$) and rotational (roll, pitch, yaw) component respectively along the $\mathbf{x}$, $\mathbf{y}$, and $\mathbf{z}$ axis. Note that unlike most 2D navigation problems in which the next vehicle state $\mathbf{x}_{t+1}$ only relies on the current vehicle state $\mathbf{x}_t$ and input $\mathbf{u}_t$ alone, our next vehicle state is additionally affected by the vertically challenging terrain underneath and in front of the vehicle in the current time step, denoted as $\mathbf{m}_t$. Therefore, the forward dynamics on vertically challenging terrain can be formulated as 
\begin{equation}
    \mathbf{x}_{t+1} = f_\theta(\mathbf{x}_t, \mathbf{u}_t, \mathbf{m}_t), 
    \label{eqn::model}
\end{equation}
which is parameterized by $\theta$ and will be learned in a data-driven manner. Training data of size $N$ can be collected by driving a wheeled robot on different vertically challenging terrain and recording the current and next state, current terrain, and current input: $\mathcal{D} = \{ \langle \mathbf{x}_t, \mathbf{x}_{t+1}, \mathbf{m}_t, \mathbf{u}_t \rangle_{t=1}^N$\}. Then we learn $\theta$ by minimizing a supervised loss function: 
\begin{equation}
    \theta^* = \argmin_{\theta} \sum_{(\mathbf{x}_{t}, \mathbf{x}_{t+1}, \mathbf{m}_t, \mathbf{u}_t) \in \mathcal{D}} \lVert f_\theta(\mathbf{x}_{t}, \mathbf{u}_t, \mathbf{m}_t) - \mathbf{x}_{t+1}\rVert_H,
    \label{eqn:bc}
\end{equation}
where $||\mathbf{v}||_H = \mathbf{v}^TH\mathbf{v}$ is the norm induced by a positive definite matrix $H$, used to weigh the learning loss of the different dimensions of the vehicle state $\mathbf{x}_t$. The learned vehicle-terrain forward dynamics model $f_\theta(\cdot, \cdot, \cdot)$ can then be used to rollout future trajectories for minimal-cost planning. 

\subsection{Sampling-Based Receding-Horizon Planning}
We adopt a sampling-based receding-horizon planning paradigm, in which the planner first uniformly samples input sequences up until a short horizon $H$, uses the learned model $f_\theta$ to rollout state trajectories, evaluates their cost based on a pre-defined cost function, finds the minimal-cost trajectory, executes the first input, replans, and thus gradually moves the horizon closer to the final goal. In this way, the modeling error can be corrected by frequent replanning. However, an under-actuated wheeled robot, i.e., using $\mathbf{u}_t = (v_t, \omega_t) \in U \subset \mathbb{R}^2$ to actuate $\mathbf{x}_t = (x_t, y_t, z_t, r_t, p_t, \phi_t) \in X \subset \mathbb{SE}(3)$ subject to $f_\theta$, may easily end up in many terminal states outside of $X_\textrm{goal}$, which the vehicle cannot escape and recover from, i.e., rolling over or immobilization (getting stuck) due to excessive roll and pitch angles, irregular terrain geometry, and large height change, e.g., on a large rock. Therefore, while our goal is still to minimize the traversal time $T$ leading to $X_\textrm{goal}$, for our receding-horizon planner, we seek to optimize five cost terms on a state trajectory 
$\mathbf{x}_{0:H} = \{\textbf{x}_t\}_{t=0}^H, \textrm{s.t.}, \mathbf{x}_{t+1} = f_\theta(\mathbf{x}_t, \mathbf{u}_t, \mathbf{m}_t), \forall t<H$, 
which starts at the current time $0$ and ends at the horizon $H$, to avoid these two types of terminal states on vertical challenging terrain and also move the robot towards the goal:
\begin{equation}
\begin{split}
    c(\mathbf{x}_{0:H}) = w_1 c_{\textrm{rp}}(\mathbf{x}_{0:H}) + w_2 c_{\textrm{tg}}(\mathbf{x}_{0:H}) + w_3 c_{\textrm{hc}}(\mathbf{x}_{0:H}) \\ + w_4 c_{\textrm{mb}}(\mathbf{x}_{0:H}) + w_5 c_{\textrm{est}}(x_H),
    \label{eqn::cost}
\end{split}
\end{equation}
where $c_{\textrm{rp}}(\cdot)$, $c_{\textrm{tg}}(\cdot)$, and $c_\textrm{hc}(\cdot)$ denote the cost corresponding to the robot's (extensive) roll and pitch angle, (irregular) underneath terrain geometry, and (large) terrain height change respectively; 
$c_\textrm{mb}(\cdot)$ is the cost of moving out of the observable map boundary; 
$c_\textrm{est}(\cdot)$ is the estimated cost to reach the final goal region $X_\textrm{goal}$ from the state on the horizon $x_H$, which can be computed by the Euclidean distance $c_\textrm{est}(x_H) = ||x_H-x_G||_2$, where $x_G$ is any state inside $X_\textrm{goal}$. $w_1$ to $w_5$ are corresponding weights for the cost terms.

\subsection{Modeling Rolling-Over and Immobilization}
Vehicle roll-over is often associated with large roll and pitch angles, which we therefore seek to minimize along the state trajectory. Note that roll $r_t$ and pitch $p_t$ are part of the vehicle state $\mathbf{x}_{t}$. 
Therefore, we design a cost term that considers the absolute values of roll and pitch: 
\begin{equation}
    c_{\textrm{rp}}(\mathbf{x}_{0:H}) = w_{1,1}\sum_{t=0}^H |r_t| + w_{1,2}\sum_{t=0}^H |p_t|,
    \label{eqn::cost_ro}
\end{equation}
where $w_{1,1}$ and $w_{1, 2}$ weigh the effect of the absolute value of roll and pitch.

Similarly, vehicle immobilization often happens when the vehicle state does not change from time to time due to irregular underneath terrain geometry. Therefore, trajectories on which vehicle state significantly changes, especially along the translational dimension $x$ and $y$, are encouraged:  
\begin{equation}
    c_{\textrm{tg}}(\mathbf{x}_{0:H}) = -w_{2,1}\sum_{t=1}^H |x_t-x_{t-1}| - w_{2,2}\sum_{t=1}^H |y_t-y_{t-1}|,
     \label{eqn::cost_im}
\end{equation}
where $w_{2,1}$ and $w_{2, 2}$ weigh the effect of the displacement along $x$ and $y$ direction. 

Furthermore, the vehicle should prefer gentle slope rather than large height change to avoid immobilization, so we encourage small displacement in the $z$ direction along the trajectory: 
\begin{equation}
    c_{\textrm{hc}}(\mathbf{x}_{0:H}) = \sum_{t=1}^H |z_t - z_{t-1}|. 
     \label{eqn::cost_hc}
\end{equation}
With the cost function (Equation~(\ref{eqn::cost})) and cost terms (Equation~(\ref{eqn::cost_ro}), (\ref{eqn::cost_im}), and (\ref{eqn::cost_hc})) defined, we can use any motion planner to find the minimal-cost 6-DoF state trajectory $\mathbf{x}_{0:H}$ (see details in Section~\ref{sec::implementation}). 

\begin{figure*}[ht]
    \centering
    \includegraphics[width=1\textwidth]{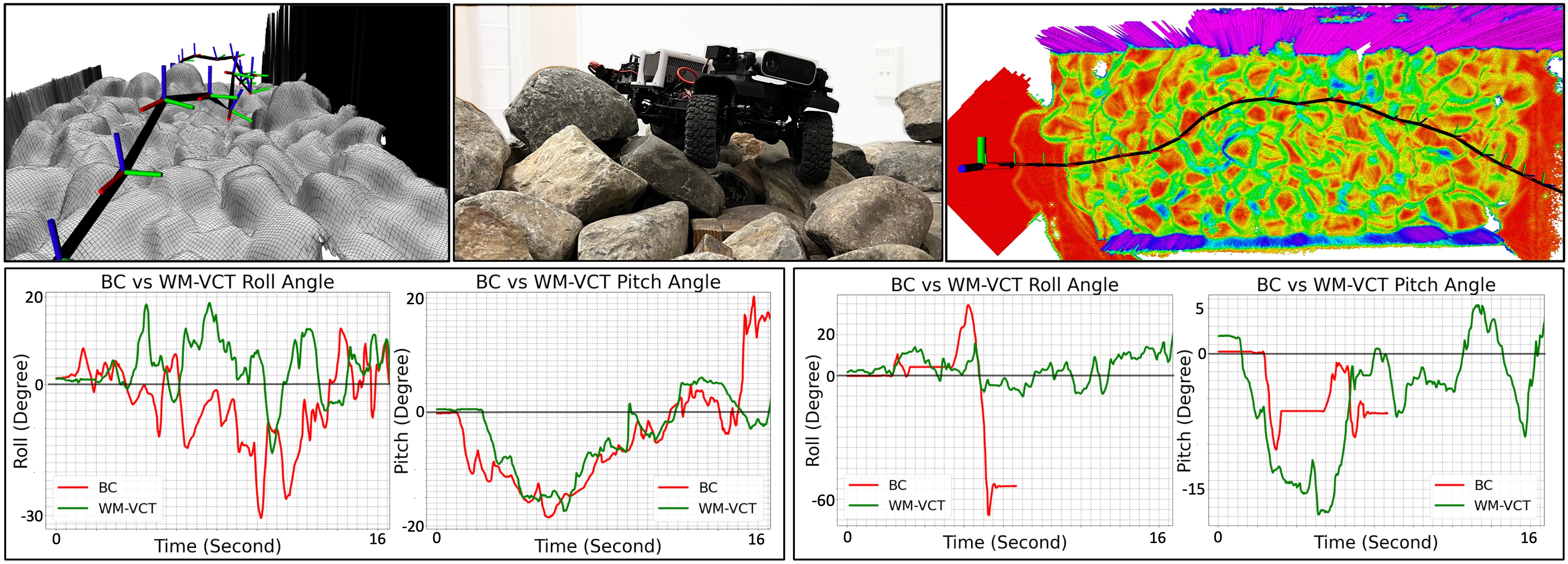}%
    \caption{Physical Experiments: The V6W (top middle) navigates through a vertical challenging environment (front and top view of the elevation map shown in top left and top right with the planned 6-DoF vehicle poses and trajectory in black); roll and pitch values of two successful \textsc{wm-vct} trials (green lines at the bottom) are shown, while BC suffers from larger values in the first (red lines bottom left) and fails the second (red lines bottom right).}
    \label{fig::experiments}
\end{figure*}

%% file: content/implementation.tex
\section{IMPLEMENTATION}
\label{sec::implementation}
We present implementation details of our \textsc{wm-vct} navigation planner onboard two physical wheeled vehicles. 

\begin{algorithm}[t]
\caption{\textsc{wm-vct} Planner} \label{alg::vm-vct}
\begin{algorithmic}[1]
\footnotesize
   \STATE \textbf{Parameters:} max iteration $I$ (10), steering sample number $N$ (11), steering range $\Omega_{\textrm{min}}$ and $\Omega_{\textrm{max}}$ ($\pm0.78\textrm{rad}$), roll-out horizon $H_{\textrm{r}}$ (5), linear velocity $V$ ($0.1\textrm{m/s}$), step size $S$ (1s), update horizon $H_{\textrm{u}}$ (3), and goal\_tolerance $G$ (0.02m)
   
   \STATE \textbf{Input:} robot pose $P_{\textrm{robot}}$ and goal pose $P_{\textrm{goal}}$   
   
   \STATE $s_1 = (x_1, y_1, z_1, r_1, p_1, \phi_1) = P_{\textrm{robot}}$, $T_{\textrm{final}} = \{s_1\}$ 
   \WHILE{max iteration  $I$ not reached}
        \FOR{$\Omega_{i}$ , $i \in [1, N]$, from range $\Omega_{\textrm{min}}$ to $\Omega_{\textrm{max}}$}
            \STATE $T_i = \{s_1\}$
            \FOR{$t \in [1, H_{\textrm{r}}]$}
                \STATE $x_{t+1}, y_{t+1}, \phi_{t+1}$ = Ackermann($x_t$, $y_t$, $\phi_t$, $V$, $\Omega_i$, S)
                \STATE $z_{t+1}$ = Elevation\_Map\_Height($x_{t+1}$, $y_{t+1}$)
                \STATE $\mathbf{m}_t$ = Elevation\_Map\_Patch($x_t$, $y_t$, $\phi_t$, $x_{t+1}$, $y_{t+1}$, $\phi_{t+1}$)
                \STATE $r_{t+1}, p_{t+1} = f_\theta(\mathbf{m}_t, r_t, p_t)$ \hfill $\triangleright$ \textit{learned roll \& pitch model}
                \STATE $s_{t+1} = (x_{t+1}, y_{t+1}, z_{t+1}, r_{t+1}, p_{t+1}, \phi_{t+1})$
                \STATE $T_i.\textrm{add}(s_{t+1})$ 
                \IF{Distance($s_{t+1}$, $P_{\textrm{goal}}$) $\leq$ $G$}
                    \STATE break \hfill $\triangleright$ \textit{goal reached, stop rollout}
                \ENDIF
                
            \ENDFOR
            \STATE $C_{i}$ = Calculate\_Cost($T_{i}$)
        \ENDFOR
	\STATE $T_{\textrm{best}} = T_{\argmin_i(C_i)}$ \hfill $\triangleright$ \textit{minimal-cost traj. up to roll-out horizon}
        \FOR{$s_j$ in $T_{\textrm{best}}$, $j\in [1, H_{\textrm{u}}]$ }
            \STATE $T_{\textrm{final}}.\textrm{add}(s_j$) $\triangleright$ \textit{backtrack to and add only up till update horizon}
        \ENDFOR
        \IF{Distance($T_{\textrm{final}}$[last], $P_{\textrm{goal}}$) $\leq$ $G$}
            \STATE Return $T_{\textrm{final}}$
        \ENDIF
	\STATE $s_1 = T_{\textrm{final}}$[last] \hfill $\triangleright$ \textit{update and restart new samples}
    \ENDWHILE
    \STATE Return $T_{\textrm{final}}$  
\end{algorithmic}
\end{algorithm}

\subsection{Physical Robots and Vertically Challenging Testbed}
We implement our \textsc{wm-vct} planner on two open-source wheeled robot platforms, the Verti-Wheelers (VWs)~\cite{datar2023toward}, one with six wheels (V6W, $0.863\textrm{m} \times 0.249\textrm{m} \times 0.2\textrm{m}$) and the other with four wheels (V4W, $0.523\textrm{m} \times 0.249\textrm{m} \times 0.2\textrm{m}$). Both robots are equipped with a Microsoft Azure Kinect RGB-D camera with a 1-DoF gimbal actuated by a servo to fixate the field of view on the terrain in front of the vehicle regardless of chassis pose. NVIDIA Jetson computers (ORIN and Xavier for V6W and V4W respectively) provide onboard computation. We use low-gear and lock both front and rear differentials to improve mobility on vertically challenging terrain. Both robots are tested on a $3.1\textrm{m} \times 1.3\textrm{m}$ rock testbed (with the highest vertical point of $X_\textrm{obs}$ reaching 0.6m) composed of hundreds of rocks and boulders of an average size of 30cm, similar to the size of the robots (Fig.~\ref{fig::overview}). The rocks and boulders on the testbed are shuffled many times during experiments. 
The vehicle state estimation for $\mathbf{x}_t$ is provided by an online Visual Inertia Odometry system provided by the \texttt{rtabmap\_ros} package~\cite{rtabmap}. 
In order to represent $\mathbf{m}_t$, we process the RGB-D input into an elevation map~\cite{mikielevation2022}, a 2D grid where each pixel ($8$mm resolution) indicates the height of the terrain at that point.

\subsection{\textsc{wm-vct} Planner Implementation}
Our planner implementation is shown in Algorithm \ref{alg::vm-vct}. 
\subsubsection{Dynamics Model Decomposition} 
\begin{figure}
  \centering
  \vspace{3pt}
  \includegraphics[width=0.6\columnwidth]{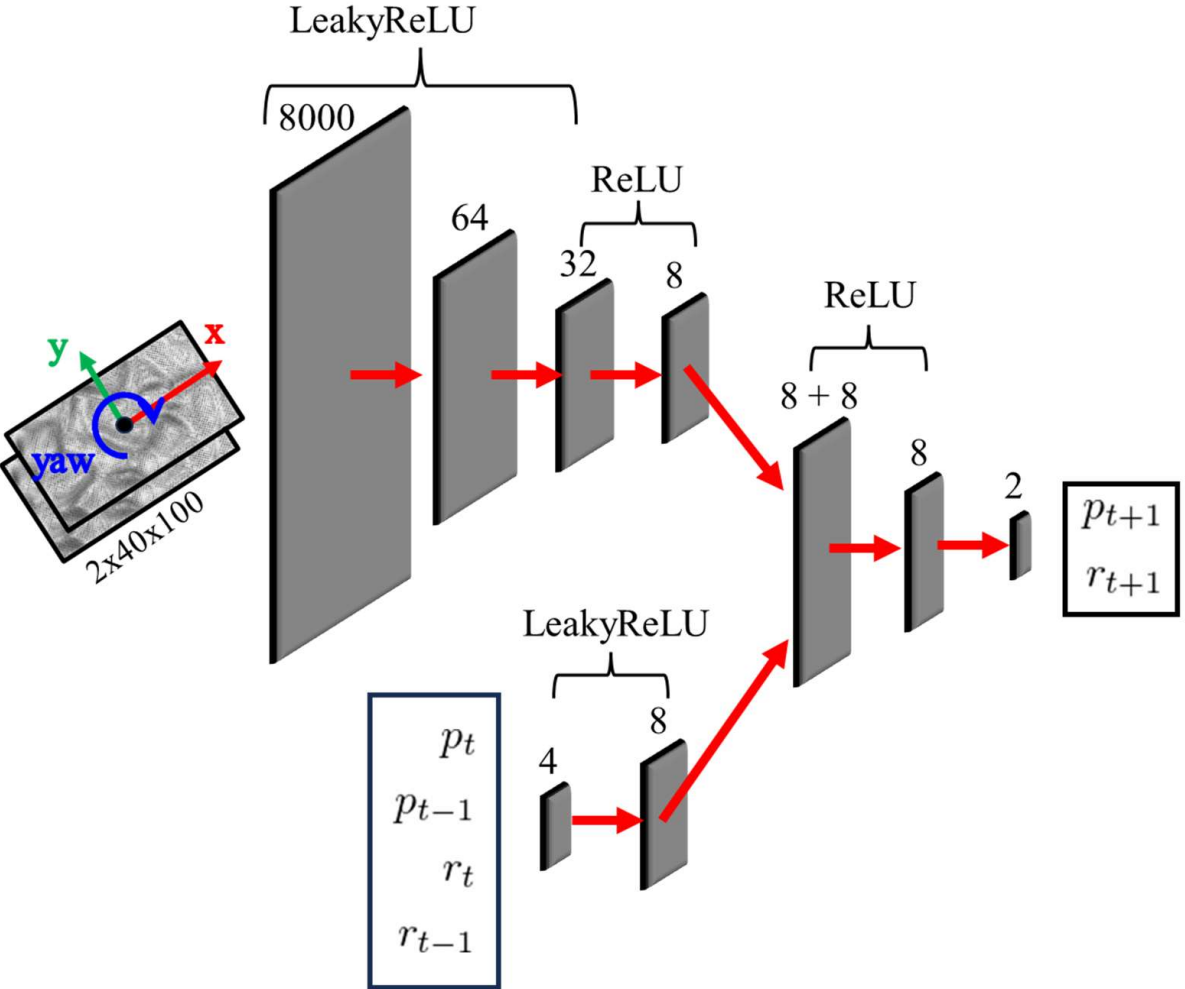}
  \caption{Neural Network Architecture for Roll and Pitch.}
  \label{fig::nn}
\end{figure}

We decompose the full 6-DoF vehicle-terrain dynamics model $\mathbf{x}_{t+1}=(x_{t+1}, y_{t+1}, z_{t+1}, r_{t+1}, p_{t+1}, \phi_{t+1}) = f_\theta(\mathbf{x}_t, \mathbf{u}_t, \mathbf{m}_t)$ into three parts: We utilize a planar Ackermann-steering model to obtain approximate $(x_{t+1}, y_{t+1}, \phi_{t+1})$ based on $(x_t, y_t, \phi_t)$ and $\mathbf{u}_t$ (line 8 in Algorithm \ref{alg::vm-vct}); $z_{t+1}$ is determined by the value of the elevation map $\mathbf{m}_t$ at $(x_{t+1}, y_{t+1})$ (line 9); We use a neural network to predict the roll $r_{t+1}$ and pitch $p_{t+1}$ angles based on $\mathbf{m}_t$, $r_t$, and $p_t$ (line 11, potentially with additional history values). In practice, we find such a decomposition very efficient and sufficiently accurate in capturing the 6-DoF vehicle dynamics with a very small amount of training data (approximately 30 minutes) and leave learning the full dynamics model as future work. 
We use two $40$$\times$$100$ ($0.32$m$\times0.8$m) elevation maps centered at both the robot's current and next position and aligned with the current and next yaw angle, i.e., underneath $(x_t, y_t)$/$(x_{t+1}, y_{t+1})$ and aligned with $\phi_t$/$\phi_{t+1}$ (line 10). 
In our neural network model (Fig.~\ref{fig::nn}), a fully connected sequential head (8000-64-32-8 neurons) processes the $2\times40\times100$ elevation maps into a 8-dimensional embedding, which is then concatenated with the 8-dimensional embedding from the last and current roll and pitch angles, $r_{t-1}$, $r_t$, $p_{t-1}$, and $p_t$,  and fed into two more fully connected layers (16-8-2), before finally producing the next roll $r_{t+1}$ and pitch $p_{t+1}$ values.
To train the model for each vehicle, we use 43161 data frames for V6W and 41367 for V4W in the open-source Verti-Wheelers datasets~\cite{datar2023toward}, roughly 30 minutes of data each demonstrating both VWs crawling over different vertically challenging testbeds. 

\subsubsection{Sampling and Roll-Out}
\begin{figure}
  \centering
  \vspace{3pt}
  \includegraphics[width=\columnwidth]{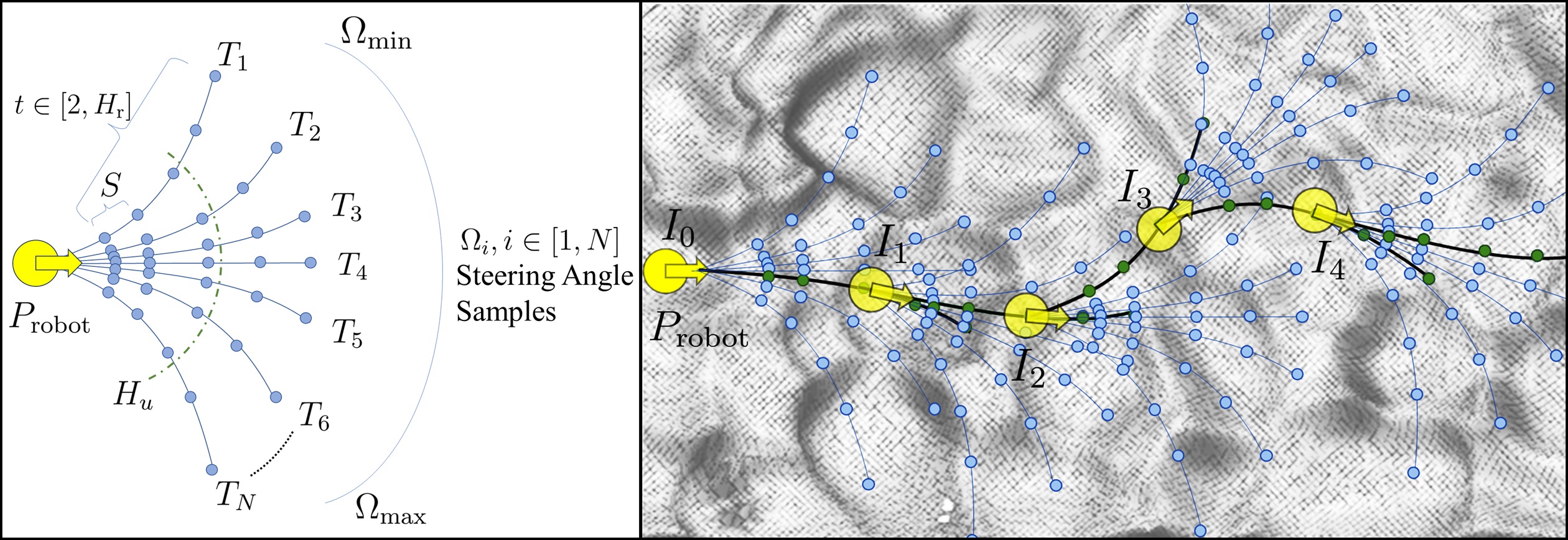}
  \caption{\textsc{wm-vct} Sampling and Roll-Out Strategy.}
  \label{fig::planner}
\end{figure}

For the sampling-based receding-horizon planner, we keep the linear velocity $v$ constant ($0.1\textrm{m/s}$) and sample 11 angular velocities $\omega$, or in our case, steering curvatures evenly from $[-0.78\textrm{rad}, 0.78\textrm{rad}]$ (line 1).
With a step size of 1 second, we roll out our vehicle dynamics model with the 11 $(v, \omega)$ pairs five times (roll-out horizon $H_{\textrm{r}}$, lines 7-17), evaluate the trajectory costs (Equation (\ref{eqn::cost}) to (\ref{eqn::cost_hc}), line 18) with corresponding cost weights  listed in Table \ref{tab::cost_function_weights}, and expand the search tree again from the 3rd state (update horizon $H_{\textrm{u}}$, lines 21-23) on the lowest-cost trajectory using the same 11 velocity pairs. We repeat this process ten times (max iteration $I$) and find the overall minimal-cost trajectory of horizon 30 ($\mathbf{x}_{0:30}$). Fig.~\ref{fig::planner} shows the \textsc{wm-vct} sampling and roll-out strategy depicted in Algorithm \ref{alg::vm-vct}. 

\begin{table}
  \caption{Cost Function Weights}
  \label{tab::cost_function_weights}
  \begin{tabular}{ccccccccc}
    \toprule
    $w_1$ & $w_2$ & $w_{1,1}$ & $w_{1,2}$ & $w_{2,1}$ & $w_{2,2}$ & $w_3$ & $w_4$ & $w_5$\\
    \midrule
    1 & 8 & 0.4 & 0.4 & 1 & 1 & 0.07 & 10 & 4\\
    \bottomrule
  \end{tabular}
\end{table}

\subsubsection{Motion Controller} 
We implement a low-level controller operating at a frequency of 30Hz, with the objective of tracking the trajectory that incurs the lowest cost. Specifically, aiming to reach the next state $\mathbf{x}_{t+1}$ from the current state $\mathbf{x}_t$, the linear velocity  controller determines the throttle command (ranging from $[-1.0, 1.0]$) by considering the current pitch angle $p_t$: the controller switches throttle command among three intervals, i.e., $0.15$ for pitch less than -5\textdegree, 0.20 for pitch from -5\textdegree\ to 5\textdegree, and 0.30 for pitch greater than 5\textdegree. Such a mechanism approximately maintains constant velocity with respect to changing terrain. The steering angle is calculated by measuring the angle between the current yaw angle $\phi_t$ and the line connecting the current 2D position $(x_t, y_t)$ to the next position $(x_{t+1}, y_{t+1})$. 
Meanwhile, while the receding-horizon planner constantly replans at 2Hz, the controller will also trigger instant replanning when the distance between the robot and the planned trajectory exceeds a predefined threshold (0.4m). 

%% file: content/experiment.tex
\section{EXPERIMENTS}
\label{sec::experiments}

\begin{table*}[ht]
\caption{Experiment Results of \textsc{bc} and \textsc{wm-vct}: Number of successful trials (out of 5), mean traversal time (of successful trials in seconds), and average roll/pitch angles (in degrees). \textsc{ol}, \textsc{rb}, and Art fail all trials. Best results are shown in bold. }
\centering
\begin{tabular}{ccccccccc}
\toprule
                   & \multicolumn{2}{c}{\footnotesize V6W}                        & \multicolumn{2}{c}{\footnotesize V4W}                        \\
\cmidrule(rl){2-3} \cmidrule(rl){4-5}
                   & \textsc{bc} & \textsc{wm-vct} & \textsc{bc} &\textsc{wm-vct} \\
\midrule
Easy      & \textbf{5}/5,  \textbf{15.8s}, 7.3\textdegree/7.9\textdegree & \textbf{5}/5,  24s, \textbf{5.1\textdegree}/\textbf{7.5\textdegree} & \textbf{2}/5,  \textbf{18.0s}, 9.2\textdegree/17.5\textdegree & \textbf{2}/5,  27.5s, \textbf{5.8\textdegree}/\textbf{9.5}\textdegree\\
Medium   & 3/5,  \textbf{17.0s}, 9.4\textdegree/\textbf{8.3}\textdegree & \textbf{4}/5,  24.5s, \textbf{6.1\textdegree}/8.6\textdegree & 1/5,  \textbf{16.0s}, 12\textdegree/\textbf{8.5}\textdegree & \textbf{2}/5,  32.5s, \textbf{7.9}\textdegree/11.4\textdegree\\
Difficult & 1/5,  \textbf{20.0s}, 8.3\textdegree/10.7\textdegree & \textbf{4}/5,  22.7s, \textbf{6.2\textdegree}/\textbf{7.4\textdegree} & N/A & N/A\\
\bottomrule
\end{tabular}
\label{tab::results}
\end{table*}

We provide experiment results and compare \textsc{wm-vct}'s performance against other baselines designed for vertically challenging terrain. 

\subsection{Baselines}

Our proposed \textsc{wm-vct} navigation planner is compared against four baselines. The three baseline algorithms developed with the open-source Verti-Wheelers project~\cite{datar2023toward}, i.e., Open-Loop (\textsc{ol}), Rule-Based (\textsc{rb}), and Behavior Cloning (\textsc{bc}), are implemented on our robots and compared against \textsc{wm-vct}. We also compare our method against the Art planner~\cite{wellhausen2023artplanner}, a state-of-the-art motion planner based on learned motion cost for quadruped robots to navigate on rough terrain, which is shown to be not applicable for our wheeled vehicle-terrain dynamics. In Fig.~\ref{fig::experiments}, we show the V6W navigating the testbed (top middle), front (top left) and top (top right) view of the elevation map with the planned 6-DoF vehicle state trajectory, and pitch and roll values in two example environments (bottom left and right). In the first environment, while both \textsc{bc} (red) and \textsc{wm-vct} (green) succeed, the former experiences larger roll and pitch values; in the second environment, \textsc{bc} (red) fails due to the excessive roll angle around 7.5s, while \textsc{wm-vct} is able to successfully navigate through. 

\subsection{Testbed Experiment Results}

We randomly shuffle the testbed three times and test all four baselines against our \textsc{wm-vct} planner. 
In all our three test environments, neither \textsc{ol} nor \textsc{rb} finish one single trial, either rolling over or getting stuck on challenging rocks. The Art planner aims at planning trajectories that go through flat surfaces for a quadruped to have stable foothold, while the quadruped torso's roll and pitch angles can be simply stabilized by the many DoFs on the four limbs, which does not apply to wheeled vehicles. It is also designed for a large quadruped robot carrying heavy-duty onboard computation and takes more than 10 seconds for one single planning cycle on the small VWs. Without timely replanning, our controller is not able to follow the outdated trajectory tailored for legged robots and therefore fails every time at difficult scenarios after deviating from the planned path. 

We present our experiment results in Table~\ref{tab::results}. The left half shows the V6W results while the right half V4W in the three obstacle courses labeled with three difficulty levels, five trials each. In general, our \textsc{wm-vct} planner achieves better results on both six-wheeled and four-wheeled platforms, compared to \textsc{bc}, the only baseline that can occasionally navigate through, in terms of navigation success rate and average roll and pitch angles. For V6W, both \textsc{bc} and \textsc{wm-vct} finish all five trials in the easy environment, with \textsc{bc} being more aggressive and therefore achieve faster traversal time but larger roll and pitch angles; for the medium difficulty, \textsc{bc} fails two trials due to rolling-over and immobilization, while \textsc{wm-vct} fails only one; in the difficult environment, \textsc{bc} succeeds only one while \textsc{wm-vct} fails only one. Note the shorter traversal time of \textsc{bc} is calculated based on only the successful trials, showing its aggressiveness, while \textsc{wm-vct} has lower roll and pitch angles in most cases (other than slightly larger pitch for Medium). V4W is a much less mechanically capable platform, and therefore fails more than V6W (it fails all trials in the difficult environment with both \textsc{bc} and \textsc{wm-vct}). But the overall comparison between \textsc{bc} and \textsc{wm-vct} remains the same: \textsc{wm-vct} finishes more trials, is slower but more stable, and achieves lower roll and pitch angles overall (except pitch for Medium). 

\subsection{Outdoor Mobility Demonstration}
\begin{figure}
  \centering
  \vspace{3pt}
  \includegraphics[width=\columnwidth]{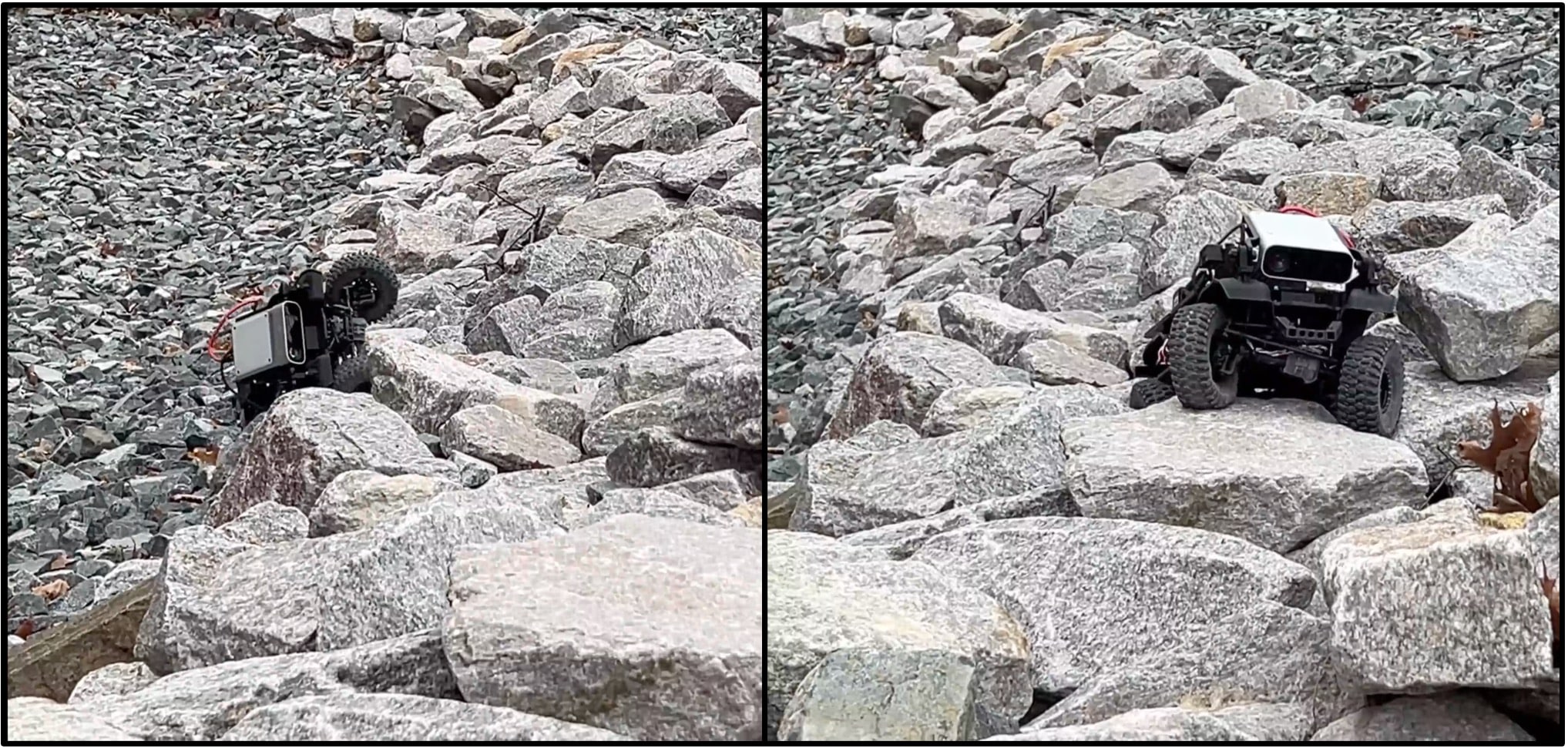}
  \caption{Outdoor Mobility Demonstration: \textsc{bc} rolls over (left) and \textsc{wm-vct} succeeds (right).}
  \label{fig::outdoor}
\end{figure}

In addition to the controlled testbed experiments, we also deploy \textsc{wm-vct} outdoors in natural vertically challenging terrain with different rock sizes to test our planner's generalizability. The outdoor environment is unseen in the training set and will challenge \textsc{bc}'s generalizability, while we expect that the limited learning scope of \textsc{wm-vct} can overcome such out-of-distribution scenarios. In fact, \textsc{wm-vct} indeed allows both V6W and V4W to avoid excessive roll and pitch angles and getting stuck on large, unseen rocks in the outdoor environment. As shown in Figure \ref{fig::outdoor}, \textsc{bc} fails because the learned end-to-end policy causes excessive roll angle and leads to roll-over (left), while \textsc{wm-vct}'s dynamic model successfully generalizes to the unseen outdoor environment. 

%% file: content/conclusion.tex
\section{CONCLUSIONS}
\label{sec::conclusions}
We present a learning approach to enable wheeled mobility on vertically challenging terrain. Going beyond the current 2D motion planning assumptions of rigid vehicle bodies and 2D planar workspaces which can be divided into free spaces and obstacles, our \textsc{wm-vct} planner first learns to model the non-rigid vehicle-terrain forward dynamics in $\mathbb{SE}(3)$ based on the current vehicle state, input, and underlying terrain. Leveraging the trajectory roll-outs under a sampling-based receding-horizon planning paradigm using the learned vehicle-terrain forward dynamics, \textsc{wm-vct} constructs a novel cost function in 3D to prevent the vehicle from rolling-over and immobilization when facing previously non-traversable obstacles. We show that our \textsc{wm-vct} planner can produce feasible, stable, and efficient motion plans to drive robots over vertically challenging terrain toward their goal and outperform several state-of-the-art baselines on two physical wheeled robot platforms. 